%% file: main.tex
\pdfoutput=1

\documentclass[11pt]{article}

\usepackage{acl}

\usepackage{times}
\usepackage{latexsym}

\usepackage[T1]{fontenc}

\usepackage[utf8]{inputenc}

\input{settings.tex}

\input{math_commands.tex}

\newcommand\our{\textsc{MetaXLM}}

\title{Cross-Lingual Language Model Meta-Pretraining}
\author{Zewen Chi,~~Heyan Huang,~~Luyang Liu,~~Yu Bai,~~Xian-Ling Mao\\
Beijing Institute of Technology \\
\texttt{\{czw,hhy63,maoxl\}@bit.edu.cn} \\}

\begin{document}

\maketitle

\begin{abstract}
The success of pretrained cross-lingual language models relies on two essential abilities, i.e., generalization ability for learning downstream tasks in a source language, and cross-lingual transferability for transferring the task knowledge to other languages. However, current methods jointly learn the two abilities in a single-phase cross-lingual pretraining process, resulting in a trade-off between generalization and cross-lingual transfer. In this paper, we propose cross-lingual language model meta-pretraining, which learns the two abilities in different training phases. Our method introduces an additional meta-pretraining phase before cross-lingual pretraining, where the model learns generalization ability on a large-scale monolingual corpus. Then, the model focuses on learning cross-lingual transfer on a multilingual corpus.
Experimental results show that our method improves both generalization and cross-lingual transfer, and produces better-aligned representations across different languages.
\end{abstract}

\section{Introduction}

Pretraining-finetuning language models (LMs) has become an emerging paradigm in natural language processing (NLP) \cite{bommasani2021opportunities}. Nonetheless, most resources like training data are English-centric, making it hard to apply this paradigm to other languages. Recently, pretrained cross-lingual LMs have shown to be effective for cross-lingual transfer, and become a promising way to build multilingual NLP applications \cite{xlm,xlmr,infoxlm}.

Cross-lingual LMs are typically pretrained with the unsupervised language modeling tasks on an unlabeled multilingual text corpus \cite{bert,xlmr,mt5}. The pretrained cross-lingual LMs achieve zero-shot cross-lingual transfer, which enables the models to be finetuned on the downstream task training data of one source language but directly applied to other target languages \cite{wu2019beto,xlingual:mbert:iclr20}.
Pretrained cross-lingual LMs can be further improved by utilizing cross-lingual supervisions from parallel corpora \cite{xlm,infoxlm}. 
Moreover, token-level alignments are also proven to be beneficial for cross-lingual transfer \cite{Cao2020Multilingual,zhao2020inducing,xlmalign}.

\begin{figure}
\centering
\includegraphics[width=0.37\textwidth]{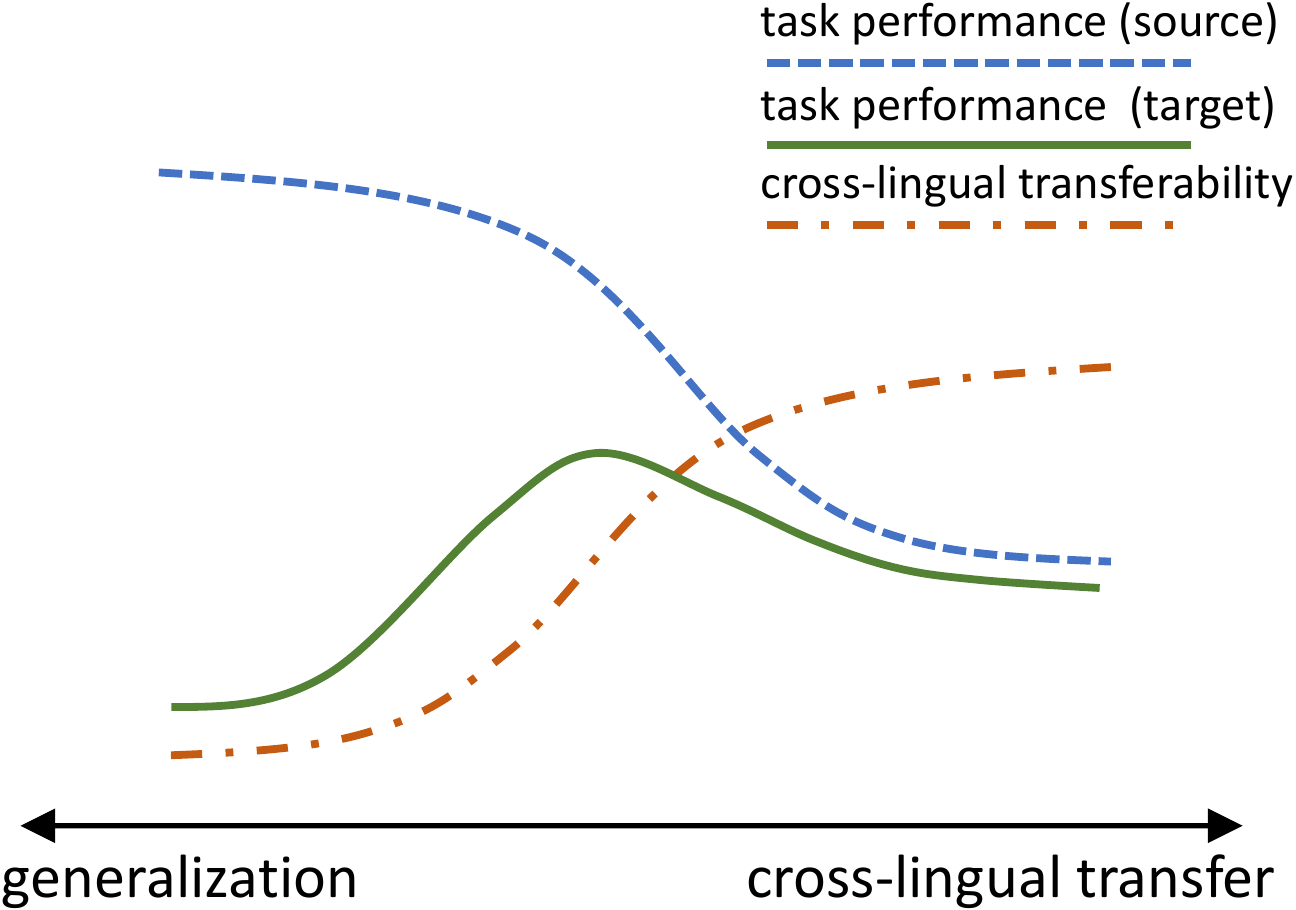}
\caption{Cross-lingual pretraining methods typically employ a single-phase cross-lingual pretraining process, resulting in  a trade-off between generalization and cross-lingual transfer.}
\label{fig:intro}
\end{figure}

The success of pretrained cross-lingual LMs relies on two essential abilities. The first ability is the generalization ability, which helps the model to learn downstream tasks in a source language. The second ability is cross-lingual transferability that transfers the downstream task knowledge to target languages.
However, current methods jointly learn the two abilities in a single-phase cross-lingual pretraining process, resulting in a trade-off between generalization and cross-lingual transfer.
The trade-off is typically achieved by a rebalanced language distribution \cite{xlm} that controls the frequency of a language to be used in cross-lingual pretraining.
As shown in Figure~\ref{fig:intro}, sampling the source language more frequently encourages generalization to downstream tasks, which provides better results on the source language. Nonetheless, it is accompanied by the weakening of cross-lingual transferability, resulting in the low performance on target languages.
On the contrary, increasing the probabilities of the target languages improves cross-lingual transfer but harms generalization.

In this paper, we propose cross-lingual language model meta-pretraining (\our{}), a novel paradigm for learning cross-lingual LMs that learns generalization ability and cross-lingual transferability in different training phases.
Our method introduces an additional meta-pretraining phase before cross-lingual pretraining, which can be understood as the pretraining of pretraining. 
Specifically, \our{} first learns generalization ability on a large-scale monolingual corpus in the meta-pretraining phase, and then focuses on learning cross-lingual transfer on a multilingual corpus in the cross-lingual pretraining phase. Finally, the model is finetuned on downstream tasks for various application scenarios of pretrained cross-lingual LMs. 

We conduct extensive experiments on ten downstream tasks under three different application scenarios, including cross-lingual understanding tasks for cross-lingual transfer, multilingual classification for supervised finetuning, and cross-lingual alignment tasks for feature-based applications. 
Experimental results show that our method provides substantial improvements over the baseline in all of the three application scenarios.
Moreover, comprehensive analysis shows that our method successfully breaks the trade-off of the single-phase pretraining, which improves both generalization and cross-lingual transfer, and produces better-aligned representations at both sentence level and word level.

Our contributions can be summarized as follows:
\begin{itemize}
\item We propose cross-lingual language model meta-pretraining, a novel paradigm for learning cross-lingual language models.
\item We conduct extensive experiments on ten downstream tasks, and demonstrate the effectiveness of our method in three application scenarios.
\item We show that our method improves both generalization and cross-lingual transfer, and produces better-aligned representations across languages, by only using additional monolingual data.
\end{itemize}

\section{Related Work}

\paragraph{Pretrained cross-lingual language models}

Cross-lingual language model pretraining is first proposed by \citet{xlm}, which aims to build universal cross-lingual encoders. They show that learning masked language modeling (MLM;~\citealt{bert}) on multilingual text improves the results on cross-lingual classification. Similarly, \textsc{mBert}~\cite{bert} learns a multilingual version of \textsc{Bert}, which is also proven to be effective for cross-lingual transfer \cite{wu2019beto,pires2019multilingual,xlingual:mbert:iclr20,xtune}.
Several studies demonstrate that pretraining cross-lingual language models at scale leads to significant performance gains \cite{xlmr,mt5}. Besides, \citet{xlm} introduce parallel corpora into cross-lingual pretraining, and show that the cross-lingual transferability can be improved by using parallel corpora during pretraining. Recent studies also present various objectives that leverage parallel corpora to improve cross-lingual LM pretraining \cite{unicoder,infoxlm,veco,hictl,nmT5,mt6,xlme}. Moreover, the word alignments implied in parallel corpora are also proven to be beneficial for cross-lingual transfer \cite{alm,Cao2020Multilingual,zhao2020inducing,hu2020explicit,xlmalign}.
Additionally, several pretrained cross-lingual models are designed for producing cross-lingual sentence embeddings \cite{tatoeba,labse}, cross-lingual natural language generation \cite{mbart,veco,mrasp,semface}, or the cross-lingual transfer of natural language generation \cite{xnlg,zmbart,deltalm}.

\begin{figure*}
\centering
\includegraphics[width=0.95\textwidth]{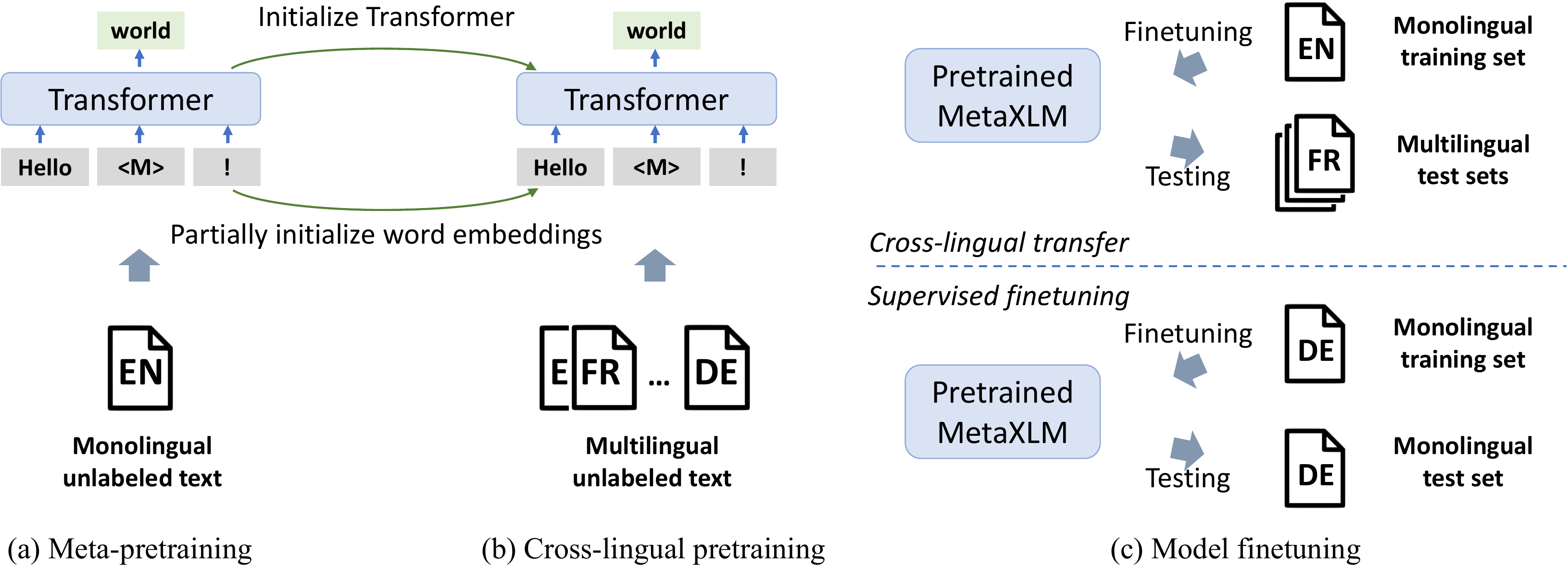}
\caption{The overview of our method, which includes three phases: (a) meta-pretraining on a monolingual corpus, (b) cross-lingual pretraining on a multilingual corpus, and (c) model finetuning on downstream tasks in different application scenarios.}
\label{fig:ov}
\end{figure*}

\paragraph{Cross-lingual transferability of monolingual models}

\citet{monotrans} demonstrate the Transformer body of a pretrained monolingual LM can also serve as a LM for another language.
Similarly, \citet{reimers2020making} propose a transferring framework that extends a monolingual model to new languages using parallel data.
\citet{li2021cross} propose a knowledge distillation method, which leverages parallel corpora to extend an existing sentence embedding model to new languages.
\citet{souza2021on} show that a pretrained monolingual LM performs well when directly finetuning it on a new language.
However, these studies focus on extending an existing monolingual model to new languages, while our goal is to pretrain a cross-lingual LM that performs cross-lingual transfer and produces universal representations. We provide related analyses in Section~\ref{sec:kd} and \ref{sec:freeze}.

\section{Method}

Figure~\ref{fig:ov} illustrates the overview of our method, which contains three phases: meta-pretraining, cross-lingual pretraining, and model finetuning. 
In this section, we first introduce the meta-pretraining phase and cross-lingual pretraining phase. Then, we present three application scenarios of cross-lingual LMs for model finetuning.

\subsection{Meta-Pretraining}
Meta-pretraining can be understood as the \textbf{pretraining of pretraining}. A pretrained model typically provides good initialization for downstream task models.
Similarly, a meta-pretrained model provides initialization for the next model pretraining phase.
In the meta-pretraining phase, we pretrain a monolingual language model on a large-scale unlabeled corpus. Following \citet{roberta}, we use a Transformer~\cite{transformer} encoder as the backbone network, and pretrain the model with the masked language modeling (MLM;~\citealt{bert}) task. The MLM task is to predict the masked words of an input text sequence. Each text sequence contains several sentences that consist of at most $512$ tokens. Note that a pretrained monolingual language model such as RoBERTa~\cite{roberta} can also be directly used as a meta-pretrained model. 

\subsection{Cross-Lingual Pretraining}

With the meta-pretrained model, the cross-lingual pretraining phase focuses on learning cross-lingual transfer. We pretrain a cross-lingual language model on a multilingual unlabeled text corpus with the MLM task. For each batch, we employ the rebalanced language sampling distribution \cite{xlm} to increase the probabilities of low-resource languages. Formally, considering a $n$-language multilingual text corpus with a size of $N_i$ for the $i$-th language, the rebalanced probability of the $i$-th language is $p_i = N_i^\alpha/(\sum_{j=1}^{n}N_j^\alpha)$, 
where $\alpha$ controls the rebalanced distribution.

We use the same model architecture with the meta-pretrained model so that the parameters of the Transformer body can be directly initialized. For the vocabulary, following \citet{xlmr}, we utilize a shared vocabulary across different languages. Thus, the resulting vocabulary is different from the meta-pretrained model. To alleviate this issue, we employ a vocabulary matching strategy. For each word in the cross-lingual vocabulary, we first directly search the word in the monolingual vocabulary, and then search the normalized word if the original one is not found.

Note that only a small part of words can be initialized because most of the words in the vocabulary are in different languages with the meta-pretrained model. An extension of the mapping strategy is to initialize more word embeddings with easily accessible bilingual dictionaries. The method is straightforward, just first searching the word translation in the dictionaries, and then applying the mapping strategy mentioned above. Using bilingual dictionaries is not strictly an unsupervised setting, but it is an intuitive way to utilize more knowledge from the meta-pretrained model.

\subsection{Model Finetuning}

The pretrained cross-lingual LM can be applied to a wide range of downstream tasks, where the word prediction layer is replaced with various task layers. Following \citet{bert}, we apply the same task layer structures with BERT for the downstream tasks. Besides, we present three application scenarios of the pretrained model as follows.

\textbf{Cross-lingual transfer} is a basic application of pretrained cross-lingual LMs. In this scenario, LMs are finetuned with training data in a source language but evaluated in various target languages. When the source language is different from the target languages, the setting is also known as zero-shot cross-lingual transfer. 

\textbf{Supervised finetuning} is a common application scenario for monolingual pretrained LMs. In this scenario, a pretrained cross-lingual LM is directly used as a monolingual model for a specific target language.

\textbf{Feature-based applications} directly use the produced contextualized sentence representations without model finetuning. For example, the extracted representations can be used for cross-lingual sentence retrieval and word alignment \cite{xtreme,simalign,xlmalign}.

\section{Experiments}

In this section, we present extensive experiments on various downstream tasks. In specific, we first evaluate the models on natural language understanding (NLU) tasks for both the cross-lingual transfer and the supervised finetuning scenarios. The two scenarios evaluate cross-lingual transfer and generalization, respectively. Then, we evaluate the models for feature-based applications on the cross-lingual alignment tasks at both sentence level and word level.

\begin{table*}[t]
\centering
\small
\scalebox{1.0}{
\begin{tabular}{l|cccccccccccccccccc}
\toprule
\multirow{2}{*}{\bf Model} & \multicolumn{2}{c}{\bf Structured (F1)} & \multicolumn{3}{c}{\bf Question Answering (F1/EM)} & \multicolumn{2}{c}{\bf Classification (Acc.)} & \multirow{2}{*}{\bf Avg} \\
& POS & NER & XQuAD & MLQA & TyDiQA & XNLI & PAWS-X & \\ \midrule
\textsc{XLM}  & 63.8 & 51.0 & 47.3 / 33.2 & 39.9 / 25.3 & 26.8 / 15.2 & 57.9 & 77.8 & 49.2 \\
\our{} &  63.7 & 54.0 & 57.9 / 42.0 & 51.5 / 34.3 & 39.7 / 23.0 & 64.6 & 83.2 & 55.7 \\
~~~~$+$Dict & \bf 67.4 & \bf 55.0 & \bf 64.4 / 49.0 & \bf 57.0 / 40.0 & \bf 44.4 / 27.8 & \bf 67.2 & \bf 84.2 & \bf 59.3 \\
\midrule
\midrule
\multicolumn{9}{l}{~~\textit{Pretraining with larger batch size and more training steps}} \\
\multicolumn{1}{l}{\textsc{mBert}~\cite{xtreme}} & 70.3 & 62.2 & 64.5 / 49.4 & 61.4 / 44.2 & 59.7 / 43.9 & 65.4 & 81.9 & 63.1 \\
\bottomrule
\end{tabular}
}
\caption{Evaluation results on XTREME cross-lingual understanding tasks under the cross-lingual transfer setting. Models are finetuned on the English training data but evaluated on all target languages. All results are averaged over five random seeds.}
\label{table:xtreme}
\end{table*}

\subsection{Setup}

\paragraph{Data}
We use multilingual raw sentences extracted from Wikipedia dumps as the training data, including $67.4$GB unlabeled text in $94$ languages. The sentences are split into subword pieces by \texttt{sentencepiece}~\cite{sentencepiece} using the vocabulary with $250$K subwords, which is provided by XLM-R~\cite{xlmr}. For bilingual dictionaries, we use the MUSE dictionaries~\cite{muse} that consist of $45$ English-centric bilingual lexicons. More details about the training data are in Appendix.

\paragraph{Training}

The backbone of \our{} is a base-size Transformer~\cite{transformer} encoder with (L=$12$, H=$768$, A=$12$, 270M parameters). For meta-pretraining, we directly use an English RoBERTa-base~\cite{roberta} model as the resulting meta-pretrained model. For cross-lingual pretraining, the model parameters are optimized with the Adam~\cite{adam} optimizer for $200$K steps with a linear scheduled learning rate of $0.0001$, and a batch size of $64$. The \our{} pretraining procedure takes about one day with an Nvidia A100 GPU. See more pretraining hyperparameters in Appendix.
 
\paragraph{Baseline}

We compare our proposed method with the single-phase cross-lingual pretraining process, i.e., pretraining the language models from scratch.
We use the unsupervised XLM~\cite{xlm} model as the baseline model, which learns the MLM task on a multilingual unlabeled text corpus. We reimplement XLM under the same pretraining setup with \our{} for a fair comparison, i.e. pretraining with the same pretraining task, batch size, steps, learning rates, random seeds, etc. It means that \our{} and XLM only differ in whether to use a meta-pretraining phase. 
Note that our models are pretrained with significantly fewer training data and training steps than the current state-of-the-art pretrained models because of the GPU limitation.

\subsection{Cross-lingual Understanding}

We evaluate \our{} on XTREME~\cite{xtreme} cross-lingual understanding tasks under the cross-lingual transfer setting. We use seven XTREME datasets that provide training sets in English, and dev/test sets in multiple languages. The datasets are
(1) the part-of-speech (POS) tagging dataset from the Universal Dependencies v2.5~\cite{udpos} treebanks, (2) the Wikiann~\cite{panx} dataset for named entity recognition, (3) the XQuAD~\cite{monotrans} question answering dataset that provides test sets in SQuAD v1.1~\cite{squad1} format, (4) the Multilingual Question Answering (MLQA;~\citealt{mlqa}) dataset that provides SQuAD-style dev and test sets, (5) the TyDiQA-GoldP~\cite{tydiqa} dataset for information-seeking question answering, (6) the XNLI~\cite{xnli} dataset for natural language inference, and (7) the PAWS-X~\cite{pawsx} dataset for paraphrase adversary.

Table~\ref{table:xtreme} presents the evaluation results on XTREME cross-lingual understanding. For each task, the presented results are first averaged over all the test languages, and then averaged over five runs with different random seeds. The detailed results on all test languages can be found in Appendix.
It can be observed that \our{} outperforms XLM on six out of seven downstream tasks, improving the average XTREME score from $49.2$ to $55.7$. It is worth mentioning that \our{} only uses more English training data than the baseline model, but it still improves the downstream task performance for not only English but also other languages. 
It demonstrates that our method successfully utilizes the pretrained monolingual LM to learn a better cross-lingual LM, which provides substantial gains on downstream cross-lingual understanding tasks.
Besides, it can be found that simply using bilingual dictionaries for initialization provides further gains over \our{}, improving the average score from $55.7$ to $59.3$.  On XNLI and PAWS-X, \our{}$+$Dict even outperforms \textsc{mBert}, which is pretrained with significantly a larger batch size and more training steps \cite{bert}.

\begin{figure}
\centering
\includegraphics[width=0.48\textwidth]{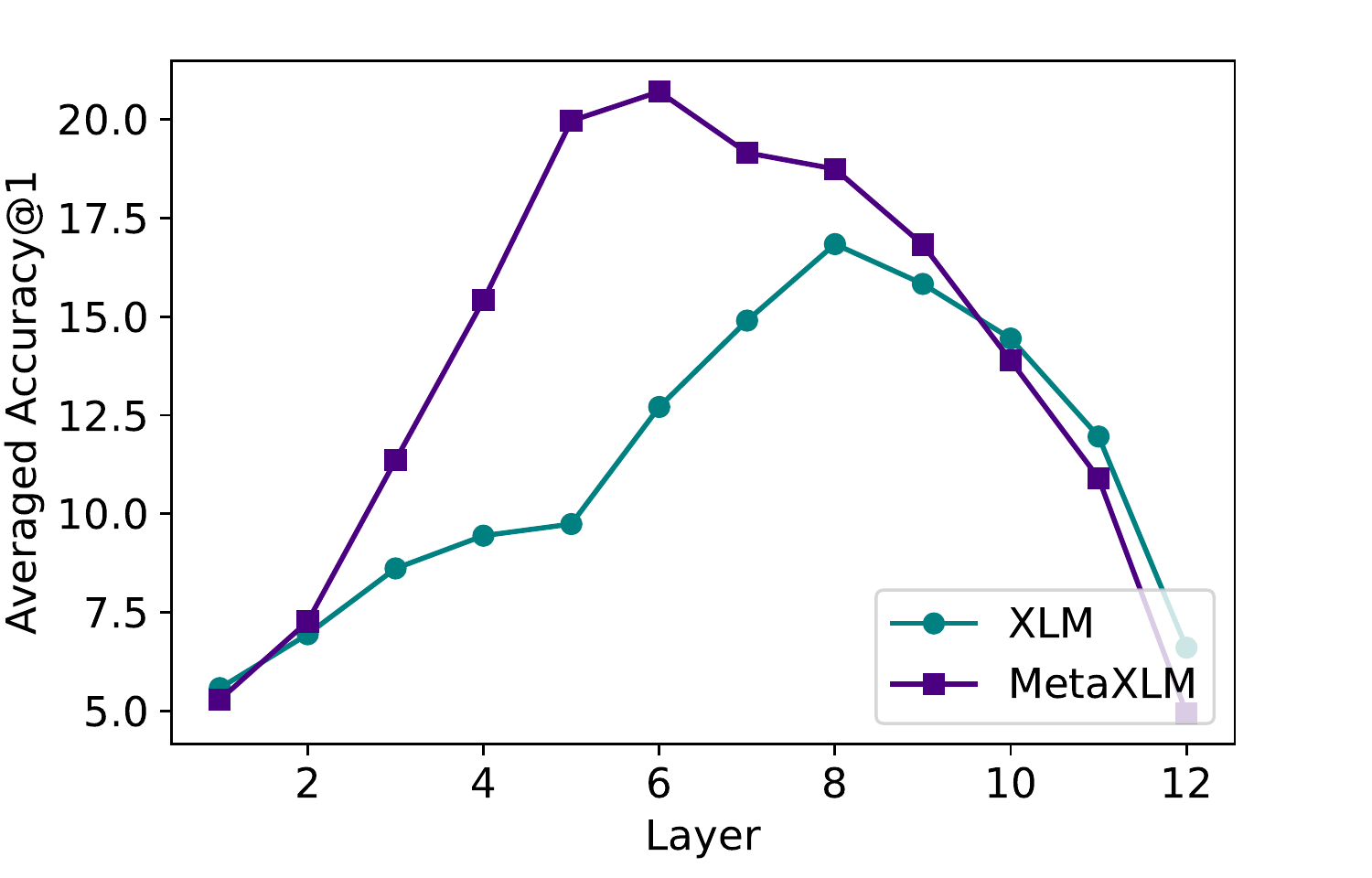}
\caption{Accuracy@1 scores on Tatoeba cross-lingual sentence retrieval. The results are averaged over 36 language pairs in both the xx $\rightarrow$ en and en $\rightarrow$ xx directions. The sentence representations are from different layers of the pretrained models.}
\label{fig:tat}
\end{figure}

\begin{table*}[t]
\centering
\small
\renewcommand\tabcolsep{5.0pt}
\scalebox{0.99}{
\begin{tabular}{l|ccc|ccc|ccc|ccc|c}
\toprule
\multirow{2}{*}{\bf Model} & \multicolumn{3}{c|}{\bf English} & \multicolumn{3}{c|}{\bf German} &  \multicolumn{3}{c|}{\bf French} &  \multicolumn{3}{c|}{\bf Japanese} & \multirow{2}{*}{\bf Avg} \\
& Book &    DVD &  Music & Book &    DVD &  Music &  Book &    DVD &  Music &  Book &    DVD &  Music &  \\ \midrule
XLM & 84.1 & 81.8 & 84.7 & 82.2 & 79.8 & 82.5 & 83.6 & 83.0 & 84.2 & 74.3 & 75.2 & 79.3 & 81.2 \\
\our{} & 89.8 & \bf 88.1 & 88.2 & \bf 86.2 & 82.0 & 84.6 & 86.1 & 86.1 & 85.8 & 76.8 & 77.5 & 80.0 & 84.3 \\
~~~~$+$Dict & \bf 90.2 & \bf 88.1 & \bf 88.6 & 85.3 & \bf 82.9 & \bf 85.8 & \bf 87.0 & \bf 86.9 & \bf 86.6 & \bf 77.8 & \bf 79.5 & \bf 80.4 & \bf 85.0 \\
\bottomrule
\end{tabular}
}
\caption{Evaluation results on Amazon Reviews multilingual classification under the supervised finetuning setting. Models are finetuned and tested under the same domain and language. All results are averaged over five random seeds. }
\label{table:cls}
\end{table*}

\subsection{Multilingual Classification}

We investigate whether our method is effective for the supervised finetuning scenario. Under this setting, the models use the same language for both finetuning and testing. Thus, this setting evaluates whether our method improves generalization for the languages other than the meta-pretraining language.
We use the multilingual Amazon Reviews dataset \cite{cls} for evaluating multilingual classification. The dataset consists of Amazon reviews written in four languages that are collected from the domains of Book, DVD, and Music. Each review is assigned with a label of `positive' or `negative'.

Table~\ref{table:cls} compares the results of \our{} with XLM. Clearly, \our{} outperforms XLM on all languages and domains, showing better generalization ability on downstream tasks than XLM. Note that our method only utilizes an additional unlabeled English corpus but still achieves better results on the other three languages. It suggests that our method successfully extracts the language-invariant knowledge from the monolingual meta-pretrained model, which improves the generalization for other languages. We also see that using bilingual dictionaries also improves the results over \our{} in the supervised finetuning scenario, which is consistent with the cross-lingual transfer scenario.

\subsection{Cross-Lingual Alignment}

To evaluate whether our method encourages the alignment of the representations from different languages, we conduct experiments on the cross-lingual sentence retrieval task and the word alignment task for cross-lingual alignment at sentence level and word level, respectively.

\paragraph{Cross-lingual sentence retrieval}
The cross-lingual sentence retrieval task has been used for evaluating the cross-lingual sentence representations produced by the pretrained cross-lingual LMs \cite{xtreme}. The goal of the task is to extract translation pairs from a bilingual document.
Following \citet{infoxlm}, we use parallel corpora in 36 English-centric language pairs from the Tatoeba~\cite{tatoeba} as the test sets. The sentences representations are the average of the hidden vectors. Then, the corresponding sentence pairs can be extracted by the nearest neighbor searching method using the cosine similarity as the distance measure. 

Figure \ref{fig:tat} reports the averaged accuracy@1 scores on Tatoeba where the sentences representations are obtained from different layers. Surprisingly, \our{} greatly improves the retrieval accuracy over the baseline model, showing that our model produces more similar sentence representations for the translation pairs than the baseline model. It indicates that our method produces universal sentence representations that are better-aligned while only using additional unlabeled English text.

\begin{figure}
\centering
\includegraphics[width=0.45\textwidth]{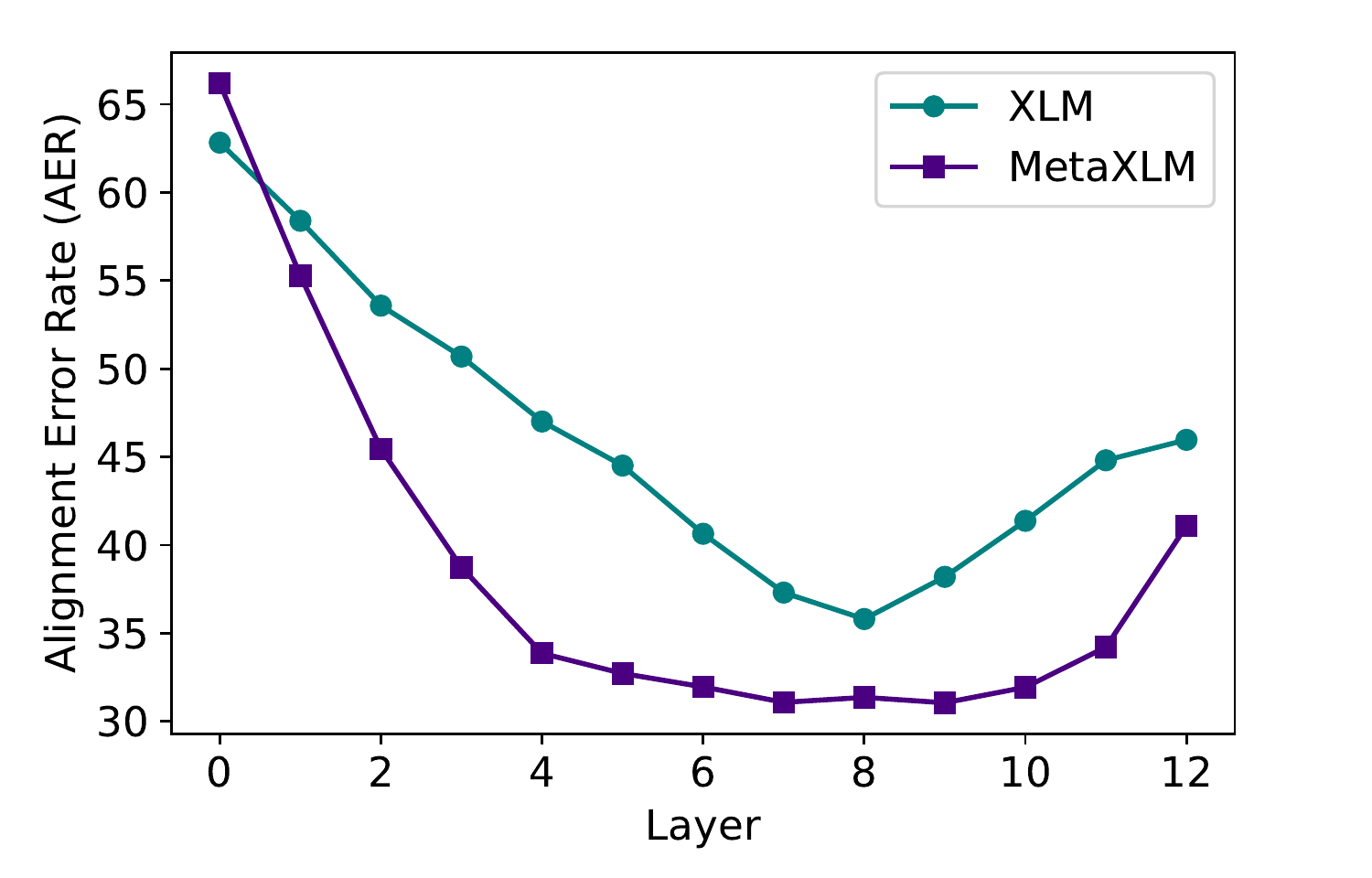}
\caption{Alignment error rates on the word alignment task, where the word representations are from different layers. The results are averaged the test sets of four language pairs.}
\label{fig:wa}
\end{figure}

\begin{figure*}
\centering
\includegraphics[width=1.0\textwidth]{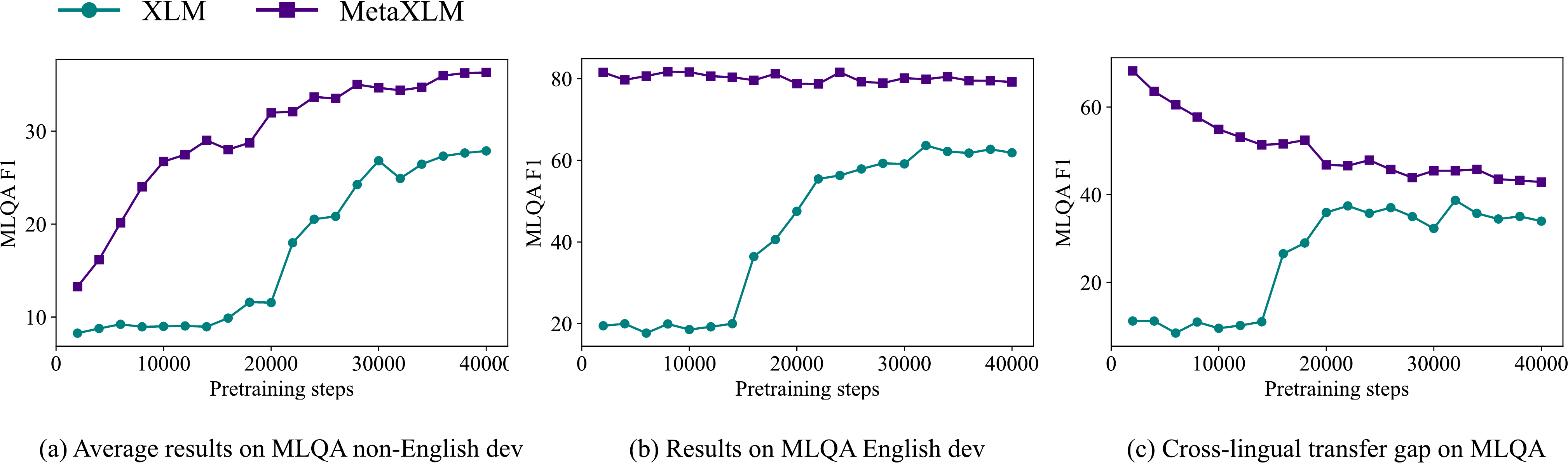}
\caption{Model performance curves along with cross-lingual pretraining. The models are evaluated on MLQA under the cross-lingual transfer setting.  (a) The F1 scores averaged over the six non-English MLQA dev sets. (b) The F1 scores on the English MLQA dev set. (c) The cross-lingual transfer gap on MLQA, i.e., the difference between the English and non-English scores. }
\label{fig:step}
\end{figure*}

\paragraph{Word alignment}

We evaluate the word-level representations with the word alignment task. The goal of the word alignment task is to extract the corresponding word pairs given an input translation pair. We use the optimal-transport alignment method~\cite{xlmalign} for the task. The method uses the hidden vectors as word representations, and produces the resulting aligned word pairs with optimal transport.
Following~\citet{simalign}, we use parallel corpora with golden alignments as test sets, which contain 1,244 translation pairs collected from EuroParl\footnote{\url{www-i6.informatik.rwth-aachen.de/goldAlignment/}}, WPT2003\footnote{\url{web.eecs.umich.edu/~mihalcea/wpt/}}, and WPT2005\footnote{\url{web.eecs.umich.edu/~mihalcea/wpt05/}}.
The resulting word pairs are evaluated by the alignment error rate (AER;~\citealt{och2003systematic}). A lower AER score indicates the word representations produces by the model are better-aligned across different languages.

Figure~\ref{fig:wa} illustrates the AER scores over different layers of the pretrained models, where layer-0 stands for the word embedding layer. It can be observed that \our{} produces the word alignments with greatly lower AER scores than XLM on all layers except the embedding layer. The results are consistent with the results of sentence retrieval, demonstrating that our model produces better-aligned representations at both sentence level and word level. Furthermore, we observe that both \our{} and XLM obtain the best performance around layer-8, while the performance of \our{} is more stable than XLM.

\section{Analysis and Discussion}

\subsection{Effects of Meta-Pretraining}

To explore how meta-pretraining improves the cross-lingual pretraining, we analyze the cross-lingual effectiveness of XLM and \our{} at various pretraining steps. We evaluate the pretrained models with the pretraining steps ranging from $2$K to $40$K on the MLQA cross-lingual question answering. Note that the models are pretrained with different steps but finetuned under the same setup.
Figure~\ref{fig:step} illustrates the model performance curves along with the cross-lingual pretraining, including (a) the F1 scores averaged over the six non-English MLQA dev sets, (b) the F1 scores on the English MLQA dev set, and (c) the cross-lingual transfer gap \cite{xtreme}. The cross-lingual transfer gap measures the difference between the score in English and the average score in the other languages. For a certain performance of English, the reduction of gap score indicates the growth of cross-lingual transferability. 

\paragraph{Lagging phenomenon of cross-lingual transfer}
We first analyze the performance curves of XLM to study the mechanism of the single-phase cross-lingual pretraining.
From Figure~\ref{fig:step} (a) and (b), we observe that the results on both English and the other languages all remain at a low level in the first $14$K steps. 
Although the performance of XLM on English has shown a rapid improvement after the $14$K step, the performance on non-English sets shows a similar trend till the $20$K step. In Figure~\ref{fig:step} (c), the transfer gap of XLM has a fast increase from the $14$K step to the $20$K step, while remaining a relatively stable transfer gap after the $20$K step. 
The above observations suggest a lagging phenomenon of cross-lingual transfer in a single-phase pretraining process, i.e., the learning of cross-lingual transfer lags behind the learning of generalization.
The observations are also consistent with \cite{dufter2020identifying}.

\paragraph{\our{} vs. XLM}
Comparing to XLM, the results of \our{} on non-English dev sets keep rising at the very beginning. Different from XLM starting with a low transfer gap, \our{} starts with a high transfer gap but continually reduces the gap during training. For the results on the English dev set, \our{} remains a high performance because of the meta-pretraining on English data. Combining the observations, it shows that our meta-pretraining method improves the cross-lingual transfer in two aspects. First, the meta-pretrained model has already obtained very good generalization ability, which enables the model to focuses on learning cross-lingual transfer. Besides, the meta-pretrained model provides good initialization for learning pretraining tasks on other languages, leading to better generalization for not only the meta-pretraining language, but also the other languages.

\begin{figure}
\centering
\includegraphics[width=0.45\textwidth]{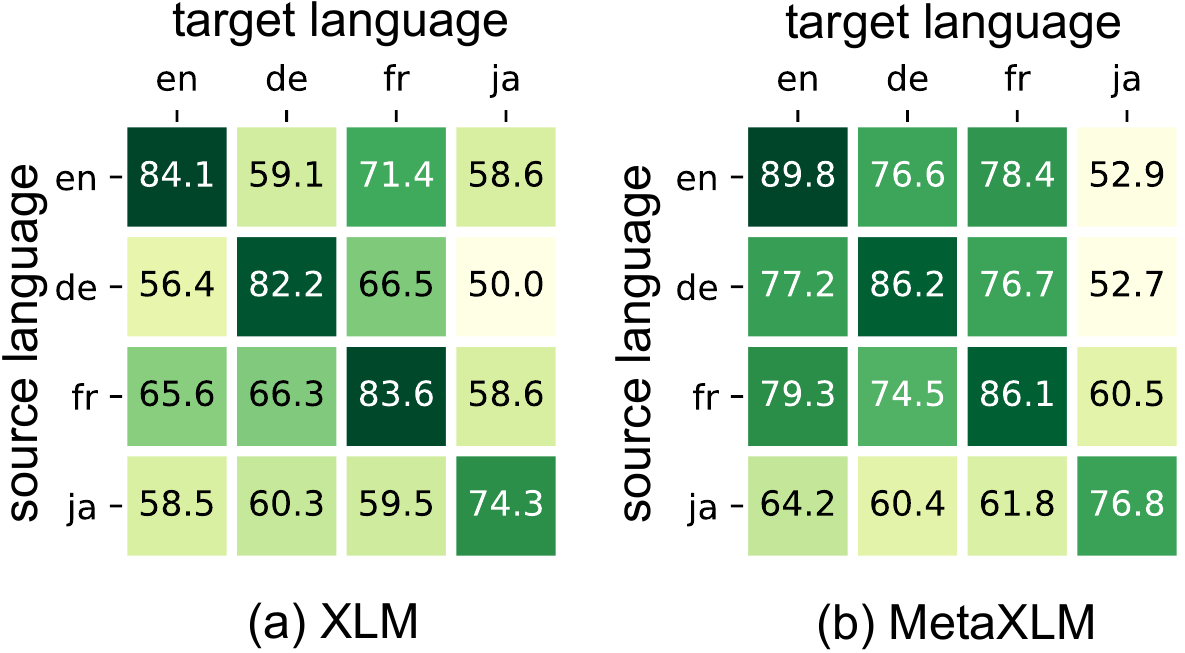}
\caption{Effects of transfer direction. The models are finetuned on Amazon Reviews using the four languages as source languages, respectively.}
\label{fig:direction}
\end{figure}

\subsection{Effects of Transfer Direction}
In addition to English-source cross-lingual transfer, we explore the cross-lingual transferability of \our{} for other transfer directions. The pretrained models are finetuned on Amazon Reviews under the cross-lingual transfer setting. Specifically, we use the four languages as the source languages respectively, and test the models on all the four languages. Figure~\ref{fig:direction} shows the results of the pretrained models for $16$ transfer directions. We see that the transferability of the pretrained models varies among different transfer directions. \our{} obtains consistent gains for $15$ out of $16$ transfer directions, indicating that using English as meta-pretraining language is beneficial for not only English-source cross-lingual transfer but also other transfer directions.
Moreover, the transfer between similar languages such as English-French can be more effective than the dissimilar directions such as German-Japanese.

\begin{table}[t]
\centering
\small
\scalebox{0.85}{
\renewcommand\tabcolsep{4.0pt}
\begin{tabular}{lccccc}
\toprule
\bf Model & \textbf{XQuAD} & \textbf{MLQA} & \textbf{TyDiQA} & \bf XNLI & \bf PAWS-X \\ \midrule
XLM & 47.3 & 39.9 & 26.8 & 57.9 & 77.8 \\
XLM$+$KD & 47.3 & 40.4 & 27.7 & 57.9 & 78.3 \\
\our{} & \bf 57.9 & \bf 51.5 & \bf 39.7 & \bf 64.6 & \bf 83.2 \\
\bottomrule
\end{tabular}
}
\caption{Effects of knowledge distillation from the monolingual model to the cross-lingual model.}
\label{table:kd}
\end{table}

\subsection{Meta-Pretraining vs. Knowledge Distillation}
\label{sec:kd}

Inspired by \citet{li2021cross} that uses knowledge distillation (KD;~\citealt{hinton2015distilling}) to improve resulting multilingual sentence embeddings.
We explore whether KD is beneficial for cross-lingual LM pretraining. We pretrain a XLM model with an auxiliary KD task, where the English RoBERTa-base~\cite{roberta} model and the cross-lingual LM are set as the teacher model, and the student model, respectively. Note that the KD task is only performed on the English examples within the batches because the teacher model is monolingual. Table~\ref{table:kd} presents the evaluation results of KD. XLM$+$KD only obtains slight improvements over XLM. Compared with \our{}, XLM$+$KD uses more computational cost, with one more network forwarding for each step, while obtaining much lower results. Thus, our method is a more practical way to utilize monolingual knowledge to improve cross-lingual pretraining.

\begin{table}[t]
\centering
\small
\scalebox{0.85}{
\renewcommand\tabcolsep{4.0pt}
\begin{tabular}{lccccc}
\toprule
\bf Model & \textbf{XQuAD} & \textbf{MLQA} & \textbf{TyDiQA} & \bf XNLI & \bf PAWS-X \\ \midrule
XLM & 47.3 & 39.9 & 26.8 & 57.9 & 77.8 \\
\our{} & \bf 57.9 & \bf 51.5 & \bf 39.7 & \bf 64.6 & \bf 83.2 \\
~~~~$+$freezing & 45.2 & 41.1 & 30.0 & 53.8 & 81.7 \\
\bottomrule
\end{tabular}
}
\caption{Effects of parameter freezing. `$+$freezing' means freezing the Transformer body during pretraining.}
\label{table:freeze}
\end{table}

\subsection{Effects of Parameter Freezing}
\label{sec:freeze}

We investigate whether freezing the initialized meta-pretrained parameters improves the cross-lingual effectiveness. Inspired by MonoTrans~\cite{monotrans}, which freezes the Transformer body to extend an English BERT to a new language, we pretrain a variant of \our{} with the Transformer body parameters frozen, i.e., only updating the embedding layers and the language modeling head during the pretraining phase.

The evaluation results are shown in Table~\ref{table:freeze}. It shows an interesting fact that \our{}$+$freezing outperforms XLM on three of the downstream tasks, while preserving a reasonable high performance on the other two tasks. It indicates that the Transformer body of the meta-pretrained monolingual LM can also serve as a Transformer body of a cross-lingual LM. This finding is also consistent with the findings in MonoTrans.
Despite the cross-lingual transferability of the Transformer body, freezing the body does not perform better than the standard \our{} on the downstream tasks. Thus, it is a better choice to update all parameters during the cross-lingual pretraining phase.

\begin{table}[t]
\centering
\small
\scalebox{0.9}{
\renewcommand\tabcolsep{5.0pt}
\begin{tabular}{lcccccc}
\toprule
\bf & \textbf{XQuAD} & \textbf{MLQA} & \textbf{TyDiQA} & \bf XNLI & \bf PAWS-X \\ \midrule
None & 47.3 & 39.9 & 26.8 & 57.9 & 77.8 \\
Body & 43.5 & 38.0 & 28.4 & 51.3 & 76.1 \\
Emb & 46.8 & 39.4 & 31.4 & 55.7 & 77.9 \\
Both & \bf 57.9 & \bf 51.5 & \bf 39.7 & \bf 64.6 & \bf 83.2 \\
\bottomrule
\end{tabular}
}
\caption{Effects of initializing different components for pretraining. `Body' and `Emb' stand for the Transformer body and the word embeddings, respectively. `None' means random initialization.}
\label{table:init}
\end{table}

\subsection{Effects of Initialization}
\label{sec:init}

We conduct experiments to study the effects of initializing different components of the model. After the meta-pretraining phase, we pretrain several variants of \our{} where we initialize the Transformer body only, the word embeddings only, or both.
Table~\ref{table:init} compares the evaluation results on cross-lingual understanding. It shows that the model produces the best results when both the Transformer body and embedding layers are initialized. 
Differently, initializing only one of the components can provide slight improvements on TyDiQA and PAWS-X, but harms the results on the other three tasks.
This suggests that the joint initialization of the two components encourages the model to extract language-invariant knowledge, which facilitates the learning of cross-lingual LMs.
On the contrary, extracting language-invariant knowledge can be difficult when only one of the components is initialized.

\section{Conclusion}

In this paper, we propose cross-lingual language model meta-pretraining, which provides a novel paradigm for learning cross-lingual language models. Extensive experiments on ten downstream tasks demonstrate the effectiveness of our method in three application scenarios. Moreover, we show that our method improves both generalization and cross-lingual transfer, and produces universal representations that are better-aligned at both sentence level and word level, by only using additional monolingual data.

Nonetheless, we should point out that our method introduces an additional meta-pretraining phase, which potentially requires more computational resources. Fortunately, pretrained monolingual models are currently easily accessible, which can be directly used as meta-pretrained models.
Besides, the meta-pretraining phase is independent of the cross-lingual pretraining phase. Thus, our method can also be easily applied to other pretrained cross-lingual models.
For future work, we would like to apply our method to language model pretraining at a larger scale. Exploring cross-lingual language model meta-pretraining for natural language generation is also an interesting research topic.

\bibliographystyle{acl_natbib}
\bibliography{main}

\newpage

\appendix

\section{Pretraining Data}

Following \citet{xlmalign}, we sample raw sentences from Wikipedia dump\footnote{\url{https://dumps.wikimedia.org/}} using a re-balanced language sampling distribution~\cite{xlm} with $\alpha=0.7$. Table~\ref{table:wiki} shows the statistics of Wikipedia dump we use.

\begin{table}[h]
\centering
\small
\scalebox{0.9}{
\begin{tabular}{crcrcr}
\toprule
Code & Size (GB) & Code & Size (GB) & Code & Size (GB) \\ \cmidrule(r){1-2}\cmidrule{3-4}\cmidrule(l){5-6}
af & 0.12 & hr & 0.28 & pa & 0.10 \\
am & 0.01 & hu & 0.80 & pl & 1.55 \\
ar & 1.29 & hy & 0.60 & ps & 0.04 \\
as & 0.04 & id & 0.52 & pt & 1.50 \\
az & 0.24 & is & 0.05 & ro & 0.42 \\
ba & 0.13 & it & 2.70 & ru & 5.63 \\
be & 0.31 & ja & 2.65 & sa & 0.04 \\
bg & 0.62 & ka & 0.37 & sd & 0.02 \\
bn & 0.41 & kk & 0.29 & si & 0.09 \\
ca & 1.10 & km & 0.12 & sk & 0.21 \\
ckb & 0.00 & kn & 0.25 & sl & 0.21 \\
cs & 0.81 & ko & 0.56 & sq & 0.11 \\
cy & 0.06 & ky & 0.10 & sr & 0.74 \\
da & 0.33 & la & 0.05 & sv & 1.70 \\
de & 5.43 & lo & 0.01 & sw & 0.03 \\
el & 0.73 & lt & 0.19 & ta & 0.46 \\
en & 12.58 & lv & 0.12 & te & 0.45 \\
eo & 0.25 & mk & 0.34 & tg & 0.04 \\
es & 3.38 & ml & 0.28 & th & 0.52 \\
et & 0.23 & mn & 0.05 & tl & 0.04 \\
eu & 0.24 & mr & 0.10 & tr & 0.43 \\
fa & 0.66 & ms & 0.20 & tt & 0.09 \\
fi & 0.68 & mt & 0.01 & ug & 0.03 \\
fr & 4.00 & my & 0.15 & uk & 2.43 \\
ga & 0.03 & ne & 0.06 & ur & 0.13 \\
gl & 0.27 & nl & 1.38 & uz & 0.06 \\
gu & 0.09 & nn & 0.13 & vi & 0.76 \\
he & 1.11 & no & 0.54 & yi & 0.02 \\
hi & 0.38 & or & 0.04 & zh & 1.08 \\
\bottomrule
\end{tabular}
}
\caption{The statistics of Wikipedia dump.}
\label{table:wiki}
\end{table}

\section{Hyperparameters}

\subsection{Cross-lingual Pretraining}

Table~\ref{table:pt-hparam} shows the hyperparameters for cross-lingual pretraining. Both the baseline model and our model are pretrained using the same hyperparameters.

\begin{table}[t]
\centering
\small
\begin{tabular}{lr}
\toprule
Hyperparameters & Value \\ \midrule
Layers & 12 \\
Hidden size & 768 \\
FFN inner hidden size & 3,072 \\
Attention heads & 12 \\
Training steps & 200K \\
Batch size & 64 \\
Adam $\epsilon$ & 1e-6 \\
Adam $\beta$ & (0.9, 0.98) \\
Learning rate & 1e-4 \\
Learning rate schedule & Linear \\
Warmup steps & 10K \\
Gradient clipping & 1.0 \\
Weight decay & 0.01 \\
Dropout rate & 0.1 \\
\bottomrule
\end{tabular}
\caption{Hyperparameters for cross-lingual pretraining.}
\label{table:pt-hparam}
\end{table}

\begin{table*}
\centering
\small
\scalebox{0.95}{
\begin{tabular}{lrrrrrrrr}
\toprule
& POS & NER & XQuAD & MLQA & TyDiQA & XNLI & PAWS-X & Amazon \\ \midrule
Batch size & 32 & 8 & 32 & 32 & 32 & 32 & 32 & 32 \\
Learning rate & 1e-5 & 7e-6 & 3e-5 & 3e-5 & 4e-5 & 1e-5 & 2e-5 & 2e-5 \\
LR schedule & Linear & Linear & Linear & Linear & Linear & Linear & Linear & Linear \\
Warmup & 10\% & 10\% & 10\% & 10\% & 10\% & 10\% & 10\% & 10\% \\
Weight decay & 0 & 0 & 0 & 0 & 0 & 0.01 & 0.01 & 0.01 \\
Epochs & 10 & 10 & 4 & 4 & 40 & 6 & 10 & 5 \\
\bottomrule
\end{tabular}
}
\caption{Hyperparameters for model finetuning.}
\label{table:hparam}
\end{table*}

\subsection{Model Finetuning}

Table~\ref{table:hparam} shows the hyperparameters for model finetuning.

\section{Detailed Results}

Table~\ref{table:udpos}-\ref{table:pawsx} show the detailed results on XTREME cross-lingual understanding tasks.

\begin{table*}[b]
\centering
\small
\renewcommand\tabcolsep{3.0pt}
\scalebox{0.8}{
\begin{tabular}{lcccccccccccc}
\toprule
 Model & en &  es &  de &  el &  ru &  tr &  ar &  vi &  th &  zh &  hi & Avg \\ \midrule
 XLM & 73.2 / 60.7 & 52.1 / 38.2 & 50.7 / 37.1 & 41.9 / 26.8 & 49.0 / 33.7 & 35.4 / 21.2 & 39.3 / 24.7 & 48.3 / 31.8 & 44.0 / 33.1 & 49.5 / 33.6 & 37.1 / 23.9 & 47.3 / 33.2 \\
\our{} & 82.0 / 70.3 & 67.0 / 53.2 & 66.4 / 52.6 & 53.1 / 34.6 & 64.4 / 47.7 & 41.6 / 24.4 & 50.8 / 33.6 & 62.2 / 42.6 & 50.9 / 37.4 & 56.0 / 38.7 & 42.2 / 27.3 & 57.9 / 42.0 \\
~~~~$+$Dict & 82.9 / 71.3 & 72.7 / 59.2 & 73.3 / 59.6 & 62.4 / 44.6 & 68.3 / 52.2 & 52.3 / 35.9 & 57.7 / 40.7 & 66.2 / 46.5 & 58.3 / 46.9 & 58.7 / 41.6 & 55.7 / 40.8 & 64.4 / 49.0 \\
\bottomrule
\end{tabular}
}
\caption{Results on XQuAD question answering.}
\label{table:xquad}
\end{table*}

\begin{table*}[b]
\centering
\small
\scalebox{0.9}{
\begin{tabular}{lcccccccc}
\toprule
 Model & en &  es &  de &  ar &  hi &  vi &  zh & Avg \\ \midrule
 XLM & 69.6 / 56.1 & 41.3 / 26.5 & 38.1 / 24.9 & 28.7 / 14.5 & 28.0 / 15.9 & 39.0 / 22.3 & 34.5 / 16.9 & 39.9 / 25.3 \\
\our{} & 79.0 / 65.5 & 55.4 / 38.2 & 52.6 / 38.2 & 41.6 / 23.3 & 36.1 / 20.3 & 53.4 / 33.8 & 42.2 / 20.7 & 51.5 / 34.3 \\
~~~~$+$Dict & 79.4 / 66.2 & 61.6 / 44.2 & 60.4 / 46.1 & 46.0 / 28.2 & 48.8 / 33.2 & 57.5 / 37.9 & 45.5 / 24.2 & 57.0 / 40.0 \\
\bottomrule
\end{tabular}
}
\caption{Results on MLQA question answering.}
\label{table:mlqa}
\end{table*}

\begin{table*}[b]
\centering
\small
\renewcommand\tabcolsep{5.0pt}
\scalebox{0.83}{
\begin{tabular}{lcccccccccc}
\toprule
 Model & en &  ar &  bn &  fi &  id &  ko &  ru &  sw &  te & Avg \\ \midrule
 XLM & 50.1 / 37.2 & 28.9 / 13.5 & 20.4 / 13.1 & 23.2 / 11.0 & 29.8 / 16.4 & 19.4 / 12.0 & 34.0 / 16.1 & 22.7 / 11.1 & 12.6 / 6.8 & 26.8 / 15.2 \\
\our{} & 67.5 / 54.2 & 48.9 / 20.4 & 27.0 / 15.2 & 43.9 / 25.2 & 50.8 / 31.2 & 23.2 / 15.0 & 51.3 / 26.9 & 30.0 / 12.3 & 15.2 / 6.8 & 39.7 / 23.0 \\
~~~~$+$Dict & 67.6 / 54.2 & 51.8 / 28.1 & 29.3 / 15.4 & 51.3 / 33.0 & 54.3 / 36.2 & 33.9 / 23.6 & 55.0 / 31.0 & 36.3 / 16.8 & 20.4 / 12.3 & 44.4 / 27.8 \\
\bottomrule
\end{tabular}
}
\caption{Results on TyDiQA question answering.}
\label{table:tydiqa}
\end{table*}

\begin{table*}[b]
\centering
\small
\scalebox{0.9}{
\begin{tabular}{lcccccccccccccccc}
\toprule
 Model & en &  fr &  es &  de &  el &  bg &  ru &  tr &  ar &  vi &  th &  zh &  hi &  sw &  ur & Avg \\ \midrule
 XLM & 74.4 & 63.4 & 64.1 & 60.4 & 59.7 & 59.0 & 58.0 & 53.0 & 55.5 & 60.2 & 54.0 & 54.5 & 55.7 & 45.3 & 51.7 & 57.9 \\
\our{} &  82.8 & 72.2 & 73.3 & 68.9 & 64.6 & 68.1 & 66.2 & 58.3 & 63.0 & 68.9 & 61.0 & 61.8 & 58.3 & 44.8 & 56.5 & 64.6 \\
~~~~$+$Dict & 83.1 & 74.2 & 75.4 & 71.4 & 68.3 & 71.7 & 70.6 & 62.8 & 65.3 & 70.2 & 65.4 & 64.5 & 63.1 & 45.7 & 56.9 & 67.2 \\
\bottomrule
\end{tabular}
}
\caption{Results on XNLI natural language inference.}
\label{table:xnli}
\end{table*}

\begin{table*}[b]
\centering
\small
\begin{tabular}{lcccccccc}
\toprule
 Model & en &  fr &  de &  es &  ja &  ko &  zh & Avg \\ \midrule
 XLM & 90.7 & 81.2 & 78.9 & 81.2 & 69.9 & 68.2 & 74.0 & 77.8 \\
\our{} & 93.8 & 88.4 & 86.2 & 87.6 & 75.5 & 72.5 & 78.2 & 83.2 \\
~~~~$+$Dict & 94.3 & 89.6 & 87.7 & 89.2 & 75.9 & 73.3 & 79.5 & 84.2 \\
\bottomrule
\end{tabular}
\caption{Results on PAWS-X paraphrase adversaries.}
\label{table:pawsx}
\end{table*}


\begin{table*}[b]
\centering
\small
\renewcommand\tabcolsep{4.0pt}
\scalebox{0.9}{
\begin{tabular}{lccccccccccccccccc}
\toprule
 Model & af &  ar &  bg &  de &  el &  en &  es &  et &  eu &  fa &  fi &  fr &  he &  hi &  hu &  id &  it \\ \midrule
XLM & 81.9 & 53.8 & 83.7 & 85.3 & 79.7 & 94.5 & 84.9 & 71.2 & 55.5 & 57.7 & 72.5 & 81.6 & 55.4 & 57.5 & 76.5 & 66.5 & 83.1 \\
\our{} & 85.6 & 61.9 & 84.8 & 84.3 & 83.2 & 95.2 & 87.3 & 72.1 & 54.1 & 65.1 & 72.5 & 84.4 & 61.1 & 47.0 & 79.4 & 70.3 & 85.5 \\
~~~~$+$Dict & 85.3 & 65.1 & 85.4 & 86.0 & 84.4 & 95.2 & 87.7 & 73.8 & 53.8 & 69.2 & 74.9 & 86.3 & 62.3 & 62.0 & 78.6 & 70.8 & 87.0 \\
\bottomrule
\end{tabular}
}
\renewcommand\tabcolsep{4.0pt}
\small
\scalebox{0.9}{
\begin{tabular}{lccccccccccccccccc}
\toprule
 Model & ja &  kk &  ko &  mr &  nl &  pt &  ru &  ta &  te &  th &  tl &  tr &  ur &  vi &  yo &  zh & Avg \\ \midrule
 XLM & 17.5 & 60.8 & 47.2 & 54.5 & 86.2 & 83.4 & 83.9 & 58.4 & 57.3 & 38.4 & 80.4 & 59.2 & 45.1 & 50.8 & 23.9 & 16.9 & 63.8 \\
\our{} & 10.8 & 60.2 & 43.6 & 51.2 & 87.4 & 86.4 & 87.3 & 53.5 & 49.1 & 33.7 & 80.1 & 58.0 & 38.7 & 53.0 & 22.0 & 14.7 & 63.7 \\
~~~~$+$Dict & 29.4 & 60.8 & 49.3 & 55.9 & 87.8 & 86.9 & 88.1 & 58.8 & 53.1 & 46.4 & 77.7 & 62.3 & 50.7 & 53.6 & 22.6 & 34.0 & 67.4 \\
\bottomrule
\end{tabular}
}
\caption{Results on part-of-speech tagging.}
\label{table:udpos}
\end{table*}

\begin{table*}[b]
\centering
\small
\renewcommand\tabcolsep{4.0pt}
\scalebox{0.8}{
\begin{tabular}{lcccccccccccccccccccc}
\toprule
 Model & ar &   he &   vi &   id &   jv &   ms &   tl &   eu &   ml &   ta &   te &   af &   nl &   en &   de &   el &   bn &   hi &   mr &   ur \\ \midrule
 XLM & 38.8 & 39.9 & 65.2 & 48.5 & 42.0 & 49.8 & 63.8 & 56.5 & 39.9 & 40.0 & 31.4 & 70.9 & 72.6 & 78.8 & 69.0 & 58.0 & 51.7 & 48.9 & 44.5 & 28.3 \\
\our{} & 35.6 & 45.0 & 64.0 & 58.2 & 45.3 & 64.3 & 66.5 & 46.4 & 42.3 & 45.6 & 35.4 & 75.4 & 79.4 & 82.7 & 74.6 & 68.5 & 57.3 & 53.2 & 42.9 & 26.1 \\
~~~~$+$Dict & 42.1 & 43.8 & 65.5 & 59.0 & 50.0 & 64.2 & 71.7 & 47.6 & 40.2 & 45.4 & 33.6 & 75.0 & 78.7 & 83.0 & 74.1 & 67.9 & 63.4 & 59.4 & 45.0 & 26.6 \\
\bottomrule
\end{tabular}
}
\renewcommand\tabcolsep{4.0pt}
\scalebox{0.8}{
\begin{tabular}{lccccccccccccccccccccc}
\toprule
 Model & fa &   fr &   it &   pt &   es &   bg &   ru &   ja &   ka &   ko &   th &   sw &   yo &   my &   zh &   kk &   tr &   et &   fi &   hu &  Avg \\ \midrule
 XLM & 32.2 & 69.9 & 72.1 & 69.2 & 61.5 & 66.3 & 49.3 & 13.6 & 53.0 & 42.3 & 0.8 & 62.8 & 42.3 & 38.5 & 16.2 & 46.7 & 62.7 & 64.0 & 70.6 & 67.5 & 51.0 \\
\our{} & 26.9 & 74.9 & 77.5 & 74.7 & 67.5 & 72.8 & 56.3 & 9.8 & 57.5 & 42.8 & 0.3 & 65.5 & 40.8 & 41.2 & 14.1 & 48.2 & 65.9 & 69.3 & 74.9 & 70.9 & 54.0 \\
~~~~$+$Dict & 28.7 & 77.4 & 78.7 & 75.0 & 72.0 & 74.1 & 57.3 & 14.3 & 54.4 & 43.1 & 0.6 & 62.7 & 41.5 & 37.5 & 17.5 & 46.2 & 66.5 & 71.6 & 74.4 & 69.4 & 55.0 \\
\bottomrule
\end{tabular}
}
\caption{Results on Wikiann named entity recognition.}
\label{table:wikiann}
\end{table*}

\end{document}

%% file: settings.tex
\usepackage{multirow}
\usepackage{amsmath}
\usepackage{capt-of}
\usepackage{tabularx}
\usepackage{epsfig}
\usepackage{amssymb}
\usepackage{amsfonts}
\usepackage{booktabs}
\usepackage{scalerel}
\usepackage[inline]{enumitem}
\usepackage{listings}
\usepackage{varwidth}
\usepackage[export]{adjustbox}
\usepackage{tikz}
\usetikzlibrary{tikzmark}
\usepackage{cleveref}

\usepackage{stmaryrd}
\usepackage{bbm}

\usepackage{algorithm}
\usepackage[noend]{algpseudocode}

\definecolor{deepblue}{rgb}{0,0,0.5}
\definecolor{officeblue}{RGB}{0,102,204}
\definecolor{deepred}{rgb}{0.6,0,0}
\definecolor{deepgreen}{rgb}{0,0.5,0}
\definecolor{mybrickred}{RGB}{182,50,28}

\definecolor{fillcolor}{RGB}{216,217,252}

\algnewcommand\algorithmicrequireb{{\hspace{0.85cm}}}
\algnewcommand\INPTDESCB{\item[\algorithmicrequireb]}

\algnewcommand\algorithmicfuncdesc{\textbf{Function:}}
\algnewcommand\FUNCDESC{\item[\algorithmicfuncdesc]}
\algnewcommand\algorithmicfuncdescb{{\hspace{1.48cm}}}
\algnewcommand\FUNCDESCB{\item[\algorithmicfuncdescb]}
\algnewcommand{\algorithmicgoto}{\textbf{goto}}
\algnewcommand{\Goto}[1]{\algorithmicgoto~\ref{#1}}



%% file: math_commands.tex
\usepackage{amsmath,amsfonts,bm}

\def\eqref#1{equation~\ref{#1}}

\def\1{\bm{1}}

\DeclareMathAlphabet{\mathsfit}{\encodingdefault}{\sfdefault}{m}{sl}
\SetMathAlphabet{\mathsfit}{bold}{\encodingdefault}{\sfdefault}{bx}{n}

